\documentclass[letterpaper, 10 pt, conference]{ieeeconf}  % Comment this line out if you need a4paper

\IEEEoverridecommandlockouts                              % This command is only needed if 
\usepackage{amsmath} % assumes amsmath package installed
\usepackage{booktabs}
\usepackage{graphicx}
\usepackage{float}
\usepackage{amssymb}
\usepackage{amsmath}
\usepackage{multirow}
\usepackage{mathrsfs}
\usepackage{caption}
\usepackage{cite}
\usepackage{cuted}

\makeatletter
\let\NAT@parse\undefined
\makeatother
\usepackage{xcolor}  
\definecolor{mypink}{RGB}{237,2,140}
\usepackage[colorlinks=true, linkcolor=blue, citecolor=mypink, urlcolor=blue]{hyperref}
\newcommand{\blue}[1]{\textcolor{black}{#1}}
\usepackage[normalem]{ulem}
\usepackage{soul}
\usepackage{xcolor} % 提供颜色功能
\definecolor{pgp}{RGB}{0, 0, 0} % 定义 pgp 为绿色
\definecolor{cwn}{RGB}{0, 0, 0} % 定义 pgp 为绿色
\newcommand{\pgp}[1]{\textcolor{pgp}{#1}}
\newcommand{\cwn}[1]{\textcolor{cwn}{#1}}
\usepackage{multirow}
\usepackage{makecell}
\usepackage[font=small,labelfont=bf]{caption}
\makeatletter
\def\@cite#1#2{[{#1\if@tempswa , #2\fi}]}

\makeatother

\title{\LARGE \bf
Trajectory-based Diffusion from One‑shot Human Video for Generalizable Manipulation
}

\author{
Guoping Pan$^{1,3,\dagger}$,
Xinmiao Yu$^{1,3,\dagger}$,
Wannan Cao$^{2,\dagger}$,
Jincheng Li$^{3}$, \\
Baolin Jiang$^{1,3}$,
Yizhe Li$^{1,3}$,
Yi Cheng$^{1,3}$,
Houde Liu$^{1,*}$
\thanks{
$^{\dagger}$These authors contributed equally to this work.
}
\thanks{
This work was supported by the Shenzhen Science and Technology Program
under Grant RCJC20210706091946001 and Grant ZDCY20250901104207008,
and by Zerith Robotics.
}
\thanks{
$^{1}$Tsinghua Shenzhen International Graduate School,
Tsinghua University, Shenzhen, China.
}
\thanks{
$^{2}$University of Science and Technology of China, Hefei, China.
}
\thanks{
$^{3}$Zerith Robotics, Shenzhen, China.
}
\thanks{
$^{*}$Corresponding author: Houde Liu.
E-mails: pgp23@mails.tsinghua.edu.cn, houdeliu@tsinghua.edu.cn
}
}
\begin{document}

\maketitle
\thispagestyle{empty}
\pagestyle{empty}

\begin{strip}
\vspace{-8em}
\begin{center}
    \captionsetup{type=figure}
    \includegraphics[width=0.98\textwidth]{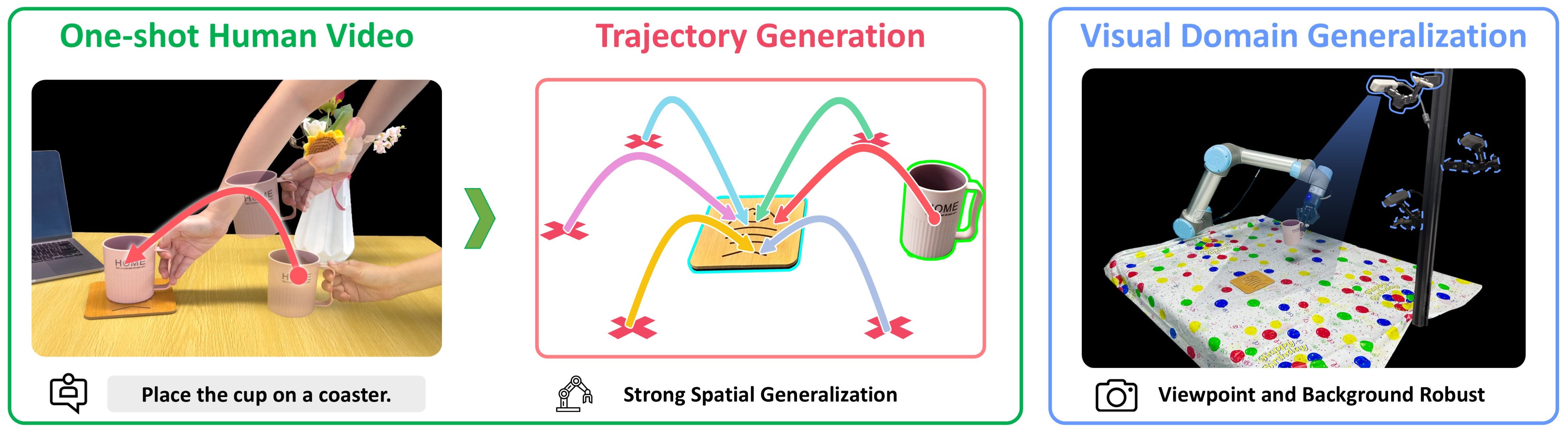}
    \captionof{figure}{
        Our method is capable of generating a large number of valid trajectories
        from a single human video, enabling strong spatial generalization.
        Moreover, it exhibits robust performance under variations in the visual
        domain during deployment.
    }
    \label{fig:surface}
\end{center}
\vspace{-1.5em}
\end{strip}

%%%%%%%%%%%%%%%%%%%%%%%%%%%%%%%%%%%%%%%%%%%%%%%%%%%%%%%%%%%%%%%%%%%%%%%%%%%%%%%%
\begin{abstract}
% Recent studies have shown that imitation learning enables robots to perform complex tasks through human demonstrations. However, common methods rely on costly teleoperation to collect data, making it difficult to scale the dataset efficiently. While human demonstration videos contain rich procedural knowledge, extracting actionable information for robot operations remains challenging due to the visual domain gap and the lack of action-related information. In this work, we introduce TDV, a trajectory-based framework that leverages data generation to enable generalizable manipulation. The key idea is to obtain an object-centered representation for each task, specifically calculating the 6D pose trajectory of the target object. This approach can effectively capture the real-time state of the object, while mitigating the effects of irrelevant background and different viewpoints. Our method outperforms previous approaches on RLBench simulation tasks, achieving an average success rate of 91.8\% across seven tasks. In real-world experiments, TDV enables cross-embodiment generalization manipulation from one-shot human demonstrations by leveraging efficient trajectory generation techniques. 
Imitation learning enables robots to perform complex tasks by leveraging human-collected demonstrations. However, most existing approaches require labor-intensive data collection with additional teleoperation systems, which hinders scalability. Human videos represent a natural source of data that implicitly contain knowledge of manipulation behaviors. However, transferring them to robots remains challenging due to the visual domain gap and the absence of explicit action labels. In this work, we introduce TDV, a trajectory-based framework that leverages data generation to enable generalizable manipulation. TDV focuses on task-relevant objects by leveraging plug-and-play 6D pose estimation techniques to extract object trajectories from human videos, thereby mitigating the impact of irrelevant backgrounds and varying viewpoints. As a modality-agnostic intermediate representation, trajectories are inherently independent of the visual domains and different embodiments, enabling efficient data generation. Our method outperforms previous approaches on RLBench simulation tasks, achieving near 100\% success rates across seven tasks. In real-world experiments, TDV enables cross-embodiment generalization manipulation from one-shot human demonstrations by leveraging efficient trajectory generation techniques.
\end{abstract}

\begin{figure*}[t]  % 用 [H] 需要 \usepackage{float}
    \centering
    \vspace{8pt}
    \includegraphics[width=0.95\linewidth]{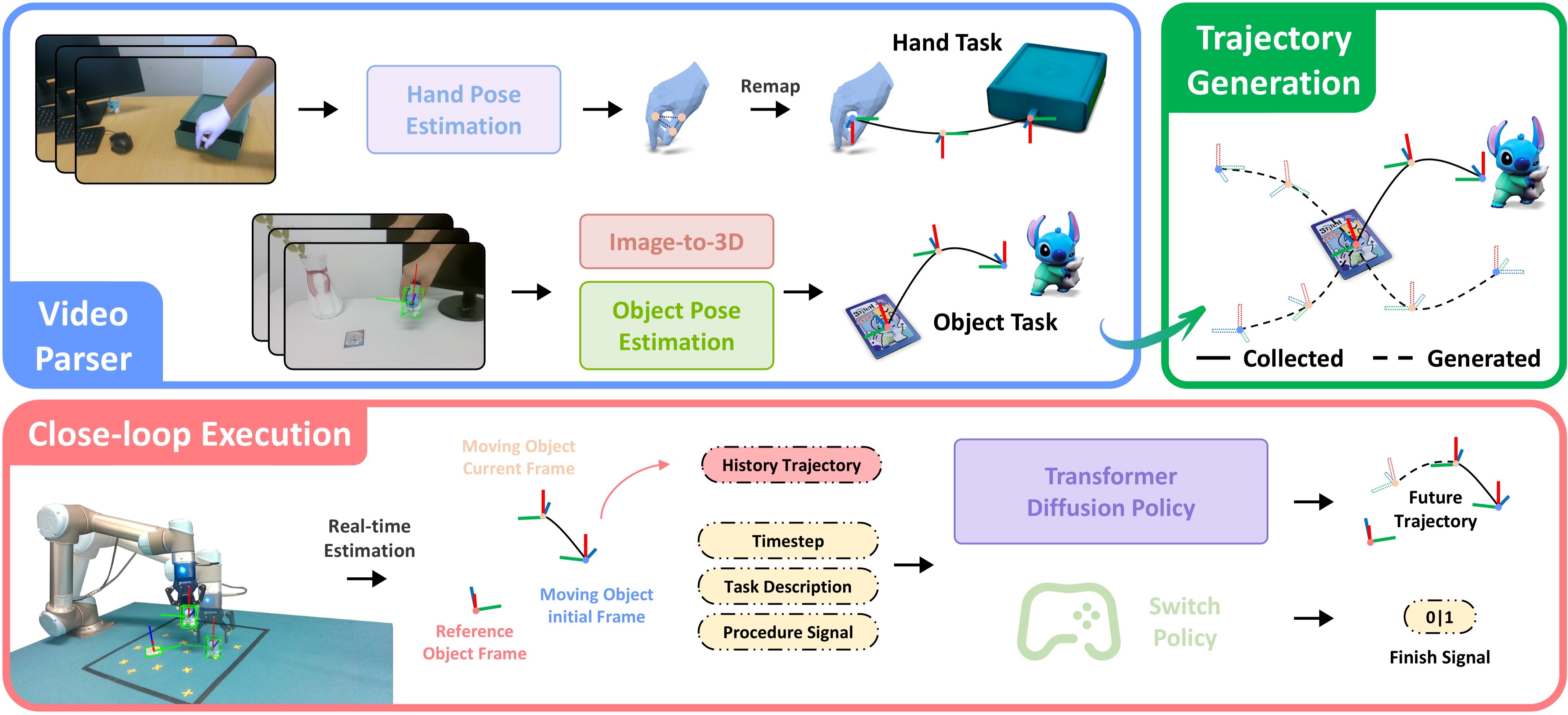}
    % \caption{Overview of TDV. 整个框架包括三个阶段。首先需要对human video进行解析，针对运动物体为手的任务通过3d hand pose estimation提取手部轨迹并进行remap到二指夹爪上。而运动物体为可操作物体时则用6d object pose estimation来提取轨迹。由于hand task运动相对固定，因此仅对object task轨迹进行生成，并用于训练trajectory-based diffusion policy. 在部署的时候通过实时检测物体位姿以预测未来轨迹和完成判断。}
    \caption{\textbf{Overview of TDV.} Our framework consists of three stages: video parsing, trajectory generation, and closed-loop execution. For hand tasks, we extract 3D hand trajectories and remap them to a two-finger gripper. For object tasks, we apply 6D object pose estimation and perform data augmentation by sampling diverse initial poses. A trajectory-based diffusion policy is trained on the resulting data. During deployment, real-time pose estimation and a switch policy enable closed-loop control and multi-procedure task execution.
    }
    \vspace{-8pt}
    \label{fig:framework}
\end{figure*}

%%%%%%%%%%%%%%%%%%%%%%%%%%%%%%%%%%%%%%%%%%%%%%%%%%%%%%%%%%%%%%%%%%%%%%%%%%%%%%%%
\section{INTRODUCTION}

Imitation learning enables robots to perform complex tasks without the need for manually configured heuristic rules. Most existing imitation learning approaches rely on proprioceptive states and visual data collected directly from the robot system, typically obtained through additional teleoperation systems operated by humans~\cite{act, DP, fu2024humanplus}. However, this approach is inefficient and presents significant challenges in scaling up data collection. In addition to images and text, there is a vast amount of human demonstration videos available on the internet, which inherently encode procedural knowledge for performing various manipulation tasks. Nevertheless, extracting meaningful perceptual information and actions from these videos for robot manipulation remains a non-trivial challenge. 

This problem involves two main aspects: the significant gap in visual domains and the lack of robot action information. \pgp{To address the visual domain gap, the most straightforward approach is to leverage image editing techniques, replacing the human with a robot in the images or erasing the human entirely~\cite{AVID, humanwild}. However, this method heavily depends on the editing quality, and there is currently a lack of studies that ensure both good consistency and compliance with physical laws.} Some studies utilize flow as an intermediate representation~\cite{flowcorl, generalflow, Gen2Act, ATM}, which captures the motion trends of task-relevant objects and exhibits robustness to visual discrepancies introduced by different embodiments. However, these methods require the camera viewpoint during deployment to be consistent with the one used during data acquisition, which limits their scalability in practical applications~\cite{xue2025demogen}. For manipulation tasks, the primary focus is on the state changes of task-relevant objects, rather than the agent executing the actions, thereby enabling cross-embodiment deployment. We find that object-centric representations, such as 6D object pose~\cite{affordance, hsu2024spot, omnimanip}, can effectively capture the real-time state of the object while mitigating the effects of irrelevant background and different viewpoints. A trajectory composed of a sequence of 6D object poses encapsulates the state transitions during the manipulation process. The previous method~\cite{hsu2024spot} requires pre-generated grasp poses for the object and fails to fully exploit the advantages of trajectory representations in scaling datasets~\cite{mandlekar2023mimicgen}, which is crucial for enhancing the generalization of manipulation.

% 我们将整个框架分为四个步骤. 首先是对human video的解析.我们根据是否存在被操作物体将操作任务划分为两种情况.当不存在操作物体时,我们将利用3D Hand Pose Estimation model对ideo中的人类手部轨迹. 当存在被操作物体时,我们利用 plug and play 的 6D pose estimation model提取它的轨迹.为了方便我们将他们称为运动物体,与之相对地另外一个物体则称之为参考物体.将参考物体的坐标作为基坐标能实现机器人基座无关.在此之后,得益于轨迹是 a modality-
% agnostic intermediate representation,我们能通过简单的轨迹缩放和插值个高效生成大量有效数据. 接着,我们利用这些数据训练一个 trajectory-based diffusion policy, 以保证generalization manipulation的鲁棒性以及multi-task manipulation conditioned on task descriptions. 最终在部署的时候我们实时监测运动物体的位姿, 将其填充到trajectory diffusion policy的历史观测队列中以生成未来的运动.
 To overcome these shortcomings, we propose a trajectory-based diffusion framework, TDV, which enables generalizable manipulation across different embodiments from one-shot human demonstrations, as shown in Fig.~\ref{fig:surface}. Our framework is structured into three stages. 
 \pgp{First, we parse human videos to extract motion trajectories relevant to the manipulation task. The full trajectory is then divided into multiple subtask segments according to changes in the moving entity or the reference object. Based on the type of moving entity, we categorize each subtask segment into two types, as illustrated in Fig.~\ref{fig:framework}: \textit{hand tasks}, where the motion is dominated by the hand itself, and \textit{object tasks}, where the motion corresponds to manipulated objects. For example, in the pouring task, grasping the kettle is treated as a hand task, while the subsequent pouring action is treated as an object task. In hand tasks, we employ a hand pose estimation model~\cite{hamer} to reconstruct human hand trajectories. For object tasks, we adopt a plug-and-play 6D pose estimation model~\cite{foundationpose} to extract object trajectories.} To eliminate the influence of viewpoint changes, we represent trajectories relative to the reference object frame. Then, the resulting trajectories provide a modality-agnostic intermediate representation, enabling efficient data augmentation through simple trajectory scaling and interpolation. Finally, we then use the augmented dataset to train a trajectory-based diffusion policy, which achieves robust generalization across diverse manipulation tasks and supports multi-procedure execution conditioned on high-level task descriptions. During deployment, the pose of the moving entity is detected in real time and fed into the history sequence of the diffusion policy, enabling closed-loop execution.
% in summary，our contribution 包括1）提出了一个仅依赖人类视频数据的trajectory-based diffusion framework, TbTd。2）根据轨迹这种modality-agnostic intermediate representation，3）实现仅依赖一条human video就可以实现空间泛化性操作。

Our contributions can be summarized as follows: 
\begin{enumerate}
    \item We propose a trajectory-based diffusion framework (TDV) that learns manipulation policies solely from human videos, significantly improving data collection efficiency and enabling scalable policy learning across diverse tasks.
    \item We leverage an object-centric intermediate representation based on trajectories to enable effective learning and cross-embodiment generalization.
    \item Our experiments demonstrate that TDV achieves strong spatial generalization and remains robust under complex backgrounds and diverse camera viewpoints.
\end{enumerate}

% Our contributions are summarized as follows: \textbf{(1)} We propose a trajectory-based diffusion framework, TDV, which learns manipulation policies solely from human videos, without requiring any robot demonstrations. \textbf{(2)} We leverage an object-centric intermediate representation based on trajectories to enable effective learning and cross-embodiment generalization. \textbf{(3)} Our experiments demonstrate that TDV achieves strong spatial generalization and remains robust under complex backgrounds and diverse camera viewpoints.
% , which enables generalizable manipulation across different embodiments from one-shot human demonstrations.
% We represent trajectories in the reference object frame to ensure their independence from the robot’s base pose.
\section{Related Works}

\subsection{Learning from Human Videos}

Many efforts have been made to learn scalable manipulation policies from non-robotic data, particularly human videos~\cite{Gen2Act,VideoDex,H-InDex,DBLP:conf/icra/ChaneSaneSL23}. The main idea is to extract meaningful representations from massive visual data and transfer them to robotic tasks. For instance, R3M~\cite{r3m} and MVP~\cite{DBLP:conf/corl/RadosavovicXJAM22} pretrain visual representations on the large-scale internet human video dataset Ego4D~\cite{ego4d}, aiming to extract manipulation knowledge beneficial for robot learning from first-person perspective videos. However, due to the heterogeneous nature of the data sources and significant domain gaps, transferring pre-trained representations to specific manipulation remains challenging. To address this issue, some methods~\cite{DBLP:conf/corl/KumarZ0J022,DBLP:conf/corl/SiebZHKF19} leverage in-domain human videos, where humans and robots perform the same manipulation tasks in identical environments. Another approach leverages transferable representations, such as affordances~\cite{vbc, humanwild} and flows~\cite{flowcorl, Gen2Act, generalflow}, combined with additional action policies.

\subsection{Object-Centric Manipulation}

To exclude background and task-irrelevant objects, many researchers focus on developing effective object-centric representations, which possess visual robustness and cross-scene generalization capabilities. The primary representation include point clouds~\cite{GROOT}, Gaussian splatting~\cite{gs}, 6D object pose estimation, neural implicit fields~\cite{NDF}, and other related methodologies~\cite{robosee, flowcorl, 3dda, hsu2024spot}. Owing to the development of visual foundation models, previous work~\cite{GROOT} has segmented the object point cloud from the scene point cloud in a modular manner and used it as input for the imitation policy, which are robust to background changes. Rather than directly using point clouds, Kerr et al.~\cite{robosee} combined 3D Gaussians with object part motion to construct a 4D Differentiable Part Model. Among object-centric representations, the 6D object pose is comparatively straightforward and efficient. Pan et al.~\cite{omnimanip} instructed the LLM to construct constraint relations based on the coordinate of task-relevant objects, and performed real-time 6D pose tracking during closed-loop execution to optimize the trajectory.

\section{Methods}
% Our overall framework consists of three stages, as illustrated in Figure~\ref{fig:framework}. To generate training data, we first extract trajectories from human demonstration videos based on task descriptions. 根据运动对象将任务分为hand tasks和object task。 由于hand task中手部相对物体的轨迹相对固定。因此仅对object task进行数据增广。我们通过 randomly sampling valid initial poses of the moving object within the workspace， 获得 a large corpus of valid training instances. All generated data are used to train a trajectory-based diffusion transformer policy. To enable autonomous execution of multi-stage tasks, we additionally train a switch policy. During deployment, we perform real-time pose detection of the moving and reference objects to achieve closed-loop control.
\pgp{Our overall framework consists of three stages, as illustrated in Fig.~\ref{fig:framework}. To generate training data, we first extract trajectories from human videos based on task descriptions. Task segments are categorized into hand tasks and object tasks based on the type of moving entity involved. Specifically, we randomly sample valid initial poses of the moving entity within the workspace to construct a large and diverse set of training instances. All generated data are then used to train a trajectory-based diffusion transformer policy. For autonomous execution of multi-procedure manipulation tasks, we introduce a switch policy that governs transitions between subtasks. During deployment, real-time pose estimation of both the moving entity and the reference objects is performed to facilitate closed-loop control.}

\subsection{Video Parser} 
% In our framework, subtask segments are categorized into hand tasks and object tasks depending on the type of moving entity. Hand tasks involve the motion of the hand itself, whereas object tasks involve the motion of manipulated objects. 

\subsubsection{Hand Pose Trajectory}

% 为了从视频数据提取出hand pose trajectory，我们采用了 一个 3d hand pose estimation model HaMeR，which 能够重建出具备21个joint pose的mano hand model。 由于我们实验采用的末端工具为二指夹爪，因此要对human手腕pose和各个手指的关节pose进行转换。考虑到手部抓握的姿势，我们选用thumb mcp joint and ip joint 以及 index mcp joint，它们位置分别表示为Pt_mcp P_t_id Pi_mcp. 我们通过这三个点可以计算出他们的映射关系T_remap
% 其计算方式如下：
% R_remap = 
% T_remap
% 其中T_fix是不同的gripper长度与人手姿势的修正关系，对于同一个robot只需要进行一次对齐即可。
% % 关键帧提取，对齐方式

To extract hand pose trajectories from video, we adopt the HaMeR~\cite{hamer} model, which reconstructs a MANO hand model with 21 articulated joints. Since our experimental setup employs a two-finger gripper as the end-effector, we remap the finger joint poses to the gripper configuration. Considering typical grasping gestures, we select the thumb MCP joint, thumb IP joint, and index MCP joint, whose positions are denoted as $P^{\text{thumb}}_{\text{MCP}}$, $P^{\text{thumb}}_{\text{IP}}$, and $P^{\text{index}}_{\text{MCP}}$, respectively. \pgp{Based on these three points, we define the remapped pose $T_{\text{remap}}$, which is aligned with the gripper end-effector pose $T_{\text{ee}}$, and its mathematical definition is given as follows.}

\begin{equation}
    T_{\text{remap}} =
    \begin{bmatrix}
    \mathbf{x} & \mathbf{y} & \mathbf{z} & \mathbf{p} \\
    0 & 0 & 0 & 1
    \end{bmatrix}
\end{equation}

\begin{align}
    \mathbf{x} &= \big(P^{\text{thumb}}_{\text{IP}} - P^{\text{thumb}}_{\text{MCP}}\big) 
                  \times \big(P^{\text{index}}_{\text{MCP}} - P^{\text{thumb}}_{\text{MCP}}\big), \nonumber \\
    \mathbf{z} &= \mathbf{x} \times 
                  \big(P^{\text{thumb}}_{\text{MCP}} - P^{\text{index}}_{\text{MCP}}\big), \label{eq:remap}\\
    \mathbf{y} &= \mathbf{z} \times \mathbf{x}, \nonumber
\end{align}

\begin{equation}
    \mathbf{p} = \tfrac{1}{2}\Big(P^{\text{thumb}}_{\text{MCP}} + P^{\text{index}}_{\text{MCP}}\Big)
\end{equation}

\subsubsection{Object Pose Trajectory}
% 这部分就是写如何从视频中获得轨迹信息
Given a close-up image of the task-relevant object, the image-to-3D generation model TRELLIS~\cite{image23d} is capable of generating high-quality 3D models resembling the object, a capability acquired through training on a large-scale dataset of diverse objects. \pgp{However, the generated 3D assets lack scale information, which we recover by estimating the global scale factor via similarity ICP against the segmented point cloud with SAM2~\cite{ravi2024sam2}.} To enable real-time object tracking in videos based on the generated object mesh, we divide the process into two steps: initial mask generation and mask tracking. We use the task description as a text prompt for GroundingSAM~\cite{groundingsam} to obtain the initial mask corresponding to the task-relevant object. Once the initial frame mask is obtained, it is used as a mask prompt for the video predictor of SAM2, enabling consistent mask propagation across subsequent frames based on through the introduction of the memory mechanism. Given the object mesh and the corresponding mask, we utilize FoundationPose~\cite{foundationpose} to estimate the 6D object pose from raw RGB-D inputs.

\subsection{Trajectory Generation.} Although directly collecting human videos as demonstrations provides an efficiency improvement over teleoperation, achieving generalizable manipulation still requires large-scale data acquisition. Compared to visual images, using trajectories as an intermediate representation offers significant advantages in terms of computational efficiency and data augmentation~\cite{mandlekar2023mimicgen}. To maintain the feasibility of the original trajectory, we only randomize the initial pose of the target object within the workspace to generate a new initial pose $\hat{T}^g_0$. Since the trajectory is referenced to the coordinate frame of the reference object, changes in the reference object pose can be equivalent to changes in the target object pose.

% 补充生成方式
% 基于起点和终点的相对性，在进行轨迹生成时，我们不妨将轨迹的终点设为定点，以原起点为坐标原点，基于所给的M(平移向量)和r（旋转角）指定新起点进行轨迹生成。
% 对于位置部分，我们首先对原始轨迹进行归一化操作，记录轨迹上每个点相对于起点的位置偏移。并计算原起点到固定终点向量和新起点到固定终点向量，利用它们在XY平面上的投影，得到轨迹的缩放因子和两个向量夹角所生成的旋转补偿，即可对归一化轨迹进行还原。再经过向量M将轨迹平移对齐新起点，便能得到新轨迹。
% 对于旋转部分，我们选择在生成轨迹末端的固定比例区域保留原始旋转方向，以此确保末端操作与原轨迹的一致性。在起点和末端固定区域间通过Slerp算法对旋转方向进行插值，得到生成轨迹的旋转部分。
% 在分别进行完位置及旋转生成后，我们将其组合成齐次变换矩阵得到完整生成轨迹。通过相邻点间距离，在首尾点不变的情况下，对轨迹点进行均匀稀释，避免采集数据时起点终点手部停留、运动过程中非匀速运动等轨迹无关因素所产生的干扰。最终得到点数一致、均匀且具备空间泛化能力的生成轨迹。
% 改论文
% 1.引用
% 2.位置变化用新符号
% 3.解释清楚四条轨迹的关系
% 4.先定义scale，然后进行归一化。再重建
% 我们会先计算出投影向量的s和r。然后对原始轨迹归一化后获得归一化轨迹，并利用s和r对归一化轨迹进行恢复，恢复所得的结果，经过平移即可得到生成轨迹。
% 将他们投影到XY平面后，通过计算这两个距离的

Smooth transition between two initial poses, $T^g_0$ and $\hat{T}^g_0$, is achieved by applying warping~\cite{peebles2023scalable} for position and spherical linear interpolation for orientation. 
For the position component ${P}$, we first compute the vectors from the original and new starting point to the fixed endpoint, respectively denoted as ${v}_o^g$, ${\hat{v}_o^g}$ and project them onto the XY-plane, resulting in the projected vectors ${v}_{o, xy}^g$ and ${\hat{v}_{o,xy}^g}$. The original trajectory is then normalized by $|{v}_{o, xy}^g|$. Using the scale factor $|{\hat{v}_{o,xy}^g}|$, we rescale the normalized trajectory, and apply the rotation of angle \( r \), which represents the angle between ${v}_{o, xy}^g$ and ${\hat{v}_{o,xy}^g}$, to obtain a generated trajectory with its starting point at the origin. Finally, we align the starting point by translating the trajectory with the vector \( \vec{M} \), which represents the offset from the original to the new starting point, yielding the final position component of generated trajectory.
 For the orientation component ${O}$, We retain the original orientation over a fixed-proportion region near the end of the trajectory, defined by a ratio parameter $\alpha \in (0, 1)$, to ensure consistency in the final operation. Between the new start orientation and this terminal region, we interpolate the orientation using the Slerp algorithm, which enables smooth orientation changes along the trajectory.

\subsection{Trajectory-based Diffusion Transformer}
To adapt to the trajectory representation, we model the middle process of the task as $f( T^{\text{target}}_{t:t+n} | T^{\text{target}}_{t-h:t}, l, s )$, where the future trajectory $T^{\text{target}}_{t:t+n}$ of length $n$ is predicted based on the history trajectory $T^{\text{target}}_{t-h:t}$ of length $h$ of the target object, conditioned on the task description $l$ and the task procedure signal $s$. We employ a Denoising Diffusion Probabilistic Model (DDPM)~\cite{ddpms} to approximate the conditional distribution $p( T^{\text{target}}_{t:t+n} | \mathbf{O}_t )$, where the observations $\mathbf{O}_t$  are composed of the history trajectory $T^{\text{target}}_{t-h:t}$, task description $l$, and procedure signal $s$. Unlike previous work, we use the historical trajectory as input rather than just the current and past poses, which is crucial for tasks that require revisiting the same poses. We decouple the encoder from the denoising process, which leads to a significant improvement in inference efficiency. The history trajectory is mapped to the latent space via a two-layer MLP. The task description $l$ is tokenized using BERT~\cite{devlin2019bert}, while the procedure signal $s$ is represented as a learnable embedding. The features are then encoded with positional information according to their modalities and fused through a transformer encoder. To better integrate the observations, we employ a DiT-block~\cite{peebles2023scalable} as the decoder backbone, with the fused features and timestep embedding serving as the conditional information. Building upon previous work, we modify the denoising process as follows:
\begin{equation}
\mathbf{T_n}^{k-1}=\alpha (\mathbf{T_n}^{k}- \gamma \varepsilon_\theta (\mathbf{O_t}, \mathbf{T_n}^{k}, k) + \mathscr{N} (0, \sigma^2 I) )
\end{equation}
where $\mathbf{T_n}$ is a simplified notation for $T^{\text{target}}_{t:t+n}$ and $\varepsilon_\theta$ is the noise prediction network with parameters $\theta$. The parameters $\alpha$, $\gamma$, and $\sigma$ of the denoising scheduler are functions of the iteration step $k$. This is the reverse process of the diffusion process, which is detailed in previous work. The loss function is modified as follows:
\begin{equation}
\mathscr{L}=\operatorname{MSE}(\varepsilon^k, \varepsilon_\theta(\mathbf{O_t}, \mathbf{T_n}^0+\varepsilon^k, k))
\end{equation}

To achieve task autonomy, we introduce a switch policy that determines whether to increment the procedure signal based on the history trajectory and task description. \blue{When a subtask is completed, the switch policy outputs a binary signal, which is used to increment the procedure control signal and to change the state of the gripper.} When the procedure signal reaches the maximum number of procedures, it indicates the completion of the task. \pgp{The switch policy is trained using subtask segments. 
For each segment, the final frame is identified as the transition point between states. 
Thus, the terminal state is labeled as 1, and all preceding states are labeled as 0.} To address data imbalance, we set the loss weight ratio to 1:100. The switch policy is implemented as a classification network built with MLPs, which is beneficial to improving inference efficiency. 
% 看情况增加cross entropy的公式

\subsection{Close-loop Execution}
% 一个完整任务将会被分成多个子任务片段，每个子任务由trajectory-based diffusion transformer policy完成执行，而sutask之间的切换则由switch policy 进行控制。In object task, to account for temporal changes in object states, we perform real-time pose estimation for both the moving object and the reference object during deployment with FoundationPose. In hand task, we utilize the end-effector pose $T_{\text{ee}}$ as the observation of moving object. 整个pipeline在能以30Hz实时执行，能应对任务过程中物体掉落等情况。
A complete task is decomposed into multiple subtask segments. Each segment is executed by the trajectory-based diffusion transformer policy, while transitions between subtasks are governed by the switch policy. In object tasks, to account for temporal changes in object states, we perform real-time pose estimation of both the moving object and the reference object during deployment using FoundationPose. In hand tasks, the moving entity pose is represented by the end-effector pose $T_{\text{ee}}$. The entire pipeline runs in real time at 30 Hz on a single NVIDIA RTX 4090 GPU, enabling robust execution even under challenging conditions such as object drops during manipulation.

%解释表格

\section{Experiments}

% We conduct comprehensive experiments in both simulation and real world to evaluate our method, aiming to answer the following questions: \textbf{(1)} How well does our method, which learns from RGB-free trajectory data, perform in diverse manipulation tasks? \textbf{(2)} How well does our method leverage cross-embodiment data and achieve efficient data collection and augmentation? \textbf{(3)} How well does our method generalize to variations of the position of the manipulated object and camera?
% 先简单说一下实验目的
We conduct a series of experiments in both simulation and real-world environments to evaluate our method, aiming to answer the following questions: 
\begin{enumerate}
    \item \label{q:traj} How effective is the trajectory as an RGB-free intermediate representation for generalizing across diverse manipulation tasks?
    \item \label{q:oneshot} Can our method achieve generalizable manipulation from one-shot human video via efficient data augmentation across embodiments?
    \item \label{q:domain} Does our method generalize to visual domain shifts, including changes in camera viewpoints and background scenes?
\end{enumerate}
% \begin{enumerate}
%     \item How effective is the trajectory as an RGB-free intermediate representation for generalizing across diverse manipulation tasks? 
%     \item Can our method achieve generalizable manipulation from one-shot human video via efficient data augmentation across embodiments? 
%     \item Does our method generalize to visual domain shifts, including changes in camera viewpoints and background scenes?
% \end{enumerate}
% \newline 
% \textbf{Q1.} How effective is the trajectory as an RGB-free intermediate representation for generalizing across diverse manipulation tasks? 
% \newline 
% \textbf{Q2.} Can our method achieve generalizable manipulation from one-shot human video via efficient data augmentation across embodiments? 
% \newline
% \textbf{Q3.} Does our method generalize to visual domain shifts, including changes in camera viewpoints and background scenes?

\begin{figure*}  % 用 [H] 需要 \usepackage{float}
    \centering
    \vspace{8pt}
    \includegraphics[width=0.95\linewidth]{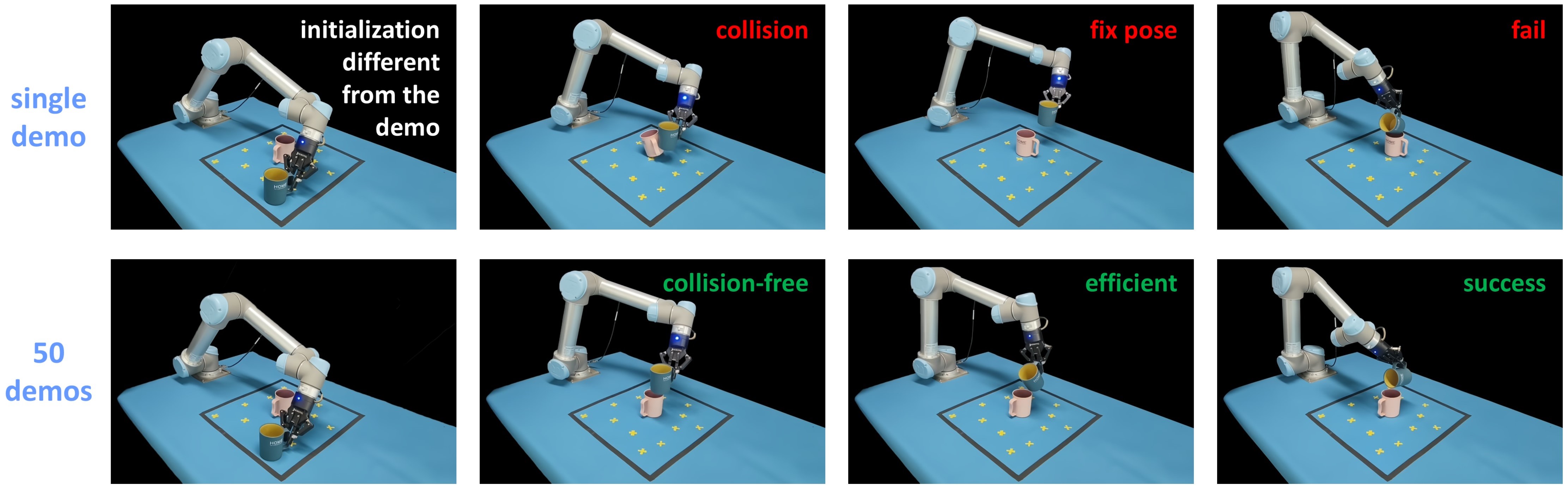}
    \caption{\textbf{Effect of data quantity on TDV generalization.} Both sequences start from an unseen test scenario. The top row, trained on a single demonstration, exhibits failure modes such as collisions and overfitting to the training trajectory. In contrast, the bottom row, trained on 50 augmented demonstrations, demonstrates reliable and efficient performance in completing the pouring task.}
    \label{fig:1vs50}
\end{figure*}

\begin{table*}[t]
    \centering
    \small
    \caption{Quantitative results on RLBench. We report success rates (\%) on 12 selected tasks in the format of mean ± standard deviation, along with the overall average across all tasks.}
    \label{table:sim_results}
    \setlength{\tabcolsep}{2.5pt}
    \renewcommand{\arraystretch}{1} % 控制行间距（1.0为默认）

    % 上半部分表格
    \begin{tabular*}{\textwidth}{@{\extracolsep{\fill}}lccccccc}
        \toprule
        % \multirow{2}{*}{\textbf{Method}} 
        % & \textbf{Close} & \textbf{Drag} & \textbf{Insert} & \textbf{Meat off} & \textbf{Place} & \textbf{Place} & \textbf{Put in} \\
        % & \textbf{Jar} & \textbf{Stick} & \textbf{Peg} & \textbf{Grill} & \textbf{Cups} & \textbf{Wine} & \textbf{Cupboard} \\
        \textbf{Method} 
            & \textbf{Close Jar} 
            & \textbf{Drag Stick} 
            & \textbf{Insert Peg} 
            & \textbf{Meat off Grill} 
            & \textbf{Place Cups} 
            & \textbf{Place Wine} 
            & \textbf{Put in Cupboard} \\
        \midrule
        OpenVLA & 1.3 ± 1.2 & 1.3 ± 1.2 & 0.0 ± 0.0 & 4.0 ± 2.0 & 2.0 ± 0.0 & 4.6 ± 1.2 & 0.0 ± 0.0 \\
        RVT2 & \textbf{100.0 ± 0.0} & \textbf{100.0 ± 0.0} & 41.3 ± 4.6 & 94.6 ± 2.3 & 40.0 ± 10.5 & 86.6 ± 2.3 & 65.3 ± 4.6 \\
        3D-DA & 49.3 ± 0.2 & \textbf{100.0 ± 0.0} & 64.0 ± 4.0 & 94.6 ± 6.1 & 22.6 ± 4.6 & 92.0 ± 4.0 & 86.6 ± 2.3 \\
        SPOT & 98.7 ± 2.3 & 80.0 ± 0.0 & 78.7 ± 2.3 & \textbf{100.0 ± 0.0} & 62.7 ± 6.1 & \textbf{100.0 ± 0.0} & 42.7 ± 4.6 \\
        \textbf{TDV (Ours)} & \textbf{100.0 ± 0.0} & \textbf{100.0 ± 0.0} & \textbf{98.7 ± 2.3} & \textbf{100.0 ± 0.0} & \textbf{73.3 ± 6.1} & 98.7 ± 2.3 & \textbf{98.7 ± 2.3} \\
        \bottomrule
    \end{tabular*}

    \vspace{0.5em} % 可选，调整上下两个表格之间的间距

    % 下半部分表格（宽度设置一致）
    \begin{tabular*}{\textwidth}{@{\extracolsep{\fill}}lccccccc}
        \toprule
        % \multirow{2}{*}{\textbf{Method}}
        % & \textbf{Put in} & \textbf{Screw} & \textbf{Sort} & \textbf{Stack} & \textbf{Stack} & \textbf{Avg.} & \\
        % & \textbf{Safe} & \textbf{Bulb} & \textbf{Shape} & \textbf{Blocks} & \textbf{Cups} & \textbf{Success} & \\
        \textbf{Method}
            & \textbf{Put in Safe}
            & \textbf{Screw Bulb}
            & \textbf{Sort Shape}
            & \textbf{Stack Blocks}
            & \textbf{Stack Cups}
            & \textbf{Avg. Success} \\
        \midrule
        OpenVLA & 4.6 ± 1.2 & 0.0 ± 0.0 & 0.0 ± 0.0 & 0.0 ± 0.0 & 0.0 ± 0.0 & 1.5 ± 0.57 & \\
        RVT2 & 96.0 ± 4.0 & \textbf{88.0 ± 0.0} & 48.0 ± 4.0 & 68.0 ± 8.0 & 76.0 ± 4.0 & 71.6 ± 3.69 & \\
        3D-DA & \textbf{100.0 ± 0.0} & 76.0 ± 0.0 & 49.3 ± 2.3 & 71.0 ± 4.0 & 53.3 ± 8.3 & 75.3 ± 2.79 & \\
        SPOT & \textbf{100.0 ± 0.0} & 48.0 ± 8.0 & 32.0 ± 4.0 & \textbf{94.0 ± 3.4} & \textbf{96.0 ± 0.0} & 79.4 ± 2.56 & \\
        \textbf{TDV (Ours)} & 98.7 ± 2.3 & 81.3 ± 2.3 & \textbf{69.3 ± 2.3} & 88.0 ± 4.0 & 94.7 ± 2.3 & \textbf{91.8 ± 2.18} & \\
        \bottomrule
    \end{tabular*}
\end{table*}

\subsection{Simulation Experiments}
\subsubsection{Benchmark}
We evaluate our method on RLBench~\cite{DBLP:journals/ral/JamesMAD20}, a standard multi-task manipulation benchmark. We select 12 representative tasks, each consisting of 2 to 60 variants, with differences in the color, shape, and target position of the objects in each variant. The selected tasks include those that require high precision and multi-step execution. Each task is specified by a language description. Following prior work~\cite{3dda}, we use 100 demonstrations per task for training and reserve 25 demonstrations for evaluation. \pgp{Our simulation experiments aim to evaluate the effectiveness of trajectories as policy observations. In this setting, trajectory extraction from human videos is not performed, and the observations of both the moving and reference objects are provided by the environment ground truth.}
% 仿真实验的设置主要目的是为了验证轨迹作为策略观测表达的有效性，因此并不涉及从人类视频中提取轨迹的过程，所采用的运动物体和参考物体观测均来自环境的groundtruth
\subsubsection{Baselines} We compare our method with several strong baselines that have demonstrated competitive performance on RLBench: 3D Diffuser Actor (3D-DA)~\cite{3dda}, RVT2~\cite{goyal2024rvt}, and SPOT~\cite{hsu2024spot}. Additionally, we include OpenVLA~\cite{kim2024openvla}, a 7B-parameter vision-language-action model specifically designed for generalization across diverse manipulation tasks. We fine-tune the released OpenVLA checkpoint using LoRA. For 3D-DA and RVT2, we follow their official training pipelines and retrain the models on our 12 selected tasks. As SPOT's implementation is not publicly available, we report its results directly from the original paper.

\subsubsection{Results}
% Quantitative results on the RLBench tasks is shown in TABLE \ref{table:sim}.We average performance across 3 random seeds with 25 trials each. Our method achieves an average success rate of 91.8\%, which is 13.4\% higher than the previous best-performing method. It achieves nearly 100\% success on most tasks. For tasks where our method's success rate is slightly lower than that of the previous method, the performance is still comparable.  Our method shows a 10.6\% improvement on long-horizon task, Place Cups, compared to the previous best, demonstrating the strong long-horizon planning and execution capabilities of our approach. Additionally, our method achieves a 20\% improvement on Sort Shape, which requires placing objects with millimeter-level accuracy into corresponding slots. These results well answer question (1) and highlight the powerful capabilities of trajectory-based methods.

Quantitative results on the RLBench tasks are presented in Table~\ref{table:sim_results}. We report the average performance over three random seeds, each with 25 trials. TDV ranks first in 8 out of 12 tasks and achieves competitive results on the remaining 4. Notably, TDV outperforms other methods by over 20\% on Insert Peg and Sort Shape, which demand high-precision manipulation. It also demonstrates more than a 10\% improvement on Place Cups and Put in Cupboard, highlighting the method's strength in long-horizon planning and execution. In tasks such as Screw Bulb and Stack Blocks, where TDV shows relatively lower performance, the main failure cases are attributed to infeasible motion planning in RLBench and errors arising from unseen relative spatial configurations. Overall, TDV achieves an average success rate of 91.8\%, which surpasses the previous state-of-the-art by 13.4\% and clearly answers Question~\ref{q:traj}.

\subsection{Real-world Experiments}

\subsubsection{Setup}
% In this section, we set up a real-world tabletop environment.We use a UR5 robot arm equipped with an adaptive gripper, and a Realsense D435 camera is fixed in front of the arm to capture global images. Five manipulation tasks are designed to evaluate our method as shown in the figure

For real-world experiments, we set up a tabletop robotic system consisting of a UR5 arm equipped with a Robotiq 2F-85 gripper. Perception is provided by an Intel RealSense D435 camera mounted on a 6-DoF articulating Magic Arm, allowing flexible adjustment of the camera viewpoint according to the experimental configuration. We evaluate our method on three object-centric and three hand-centric tasks: (1) placing the cup on a coaster (Cup-Coaster), (2) placing the Stitch toy on a card (Stitch-Card), (3) pouring water into the cup (Pour-Water), (4) pulling open the drawer (Pull-Drawer), (5) unplugging the charger from the socket (Unplug-Charger), and (6) pulling a tissue from the box (Pull-Tissue). These tasks are used to assess the generalization capabilities of our method across different object types and interaction modes.

% 在实物实验中，我们的桌面机器人系统由一个ur5和一个robotiq 2F-85夹爪组成。感知模块信息通过一个Realsense D435 camera获取，它由一个六自由度的魔术手支架进行固定，能根据实验需求方便地切换视角。为了进行实验我们分别设置了3个object tasks和hand tasks: 1) 茶杯放到杯垫上 2) 史迪奇放到卡片上 3）倒水 4）拉箱子 5）拔充电器 6）抽纸。我们通过以下一系列实验来验证我们方法的泛化性。

\begin{table}[ht]
    \centering
    \caption{Success rates under varying demonstration counts across object tasks. PG: position generalization, PoG: pose generalization.}
    \label{tab:PG}
    \begin{tabular}{c ccc ccc ccc}
        \toprule
        \multirow{2}{*}{\textbf{Num of Demo}} 
        & \multicolumn{2}{c}{\textbf{Cup-Coaster}} 
        & \multicolumn{2}{c}{\textbf{Stitch-Card}} 
        & \multicolumn{2}{c}{\textbf{Pour-Water}} \\
        \cmidrule(lr){2-3} \cmidrule(lr){4-5} \cmidrule(lr){6-7}
        & \textbf{PG} & \textbf{PoG} 
        & \textbf{PG} & \textbf{PoG} 
        & \textbf{PG} & \textbf{PoG} \\
        \midrule
        \textbf{1}   &  6/13  &  6/13  &  4/13  &  5/13  &  4/13  &  1/13  \\
        \textbf{15}  &  7/13  &  11/13  &  10/13  &  8/13  &  12/13  &  9/13  \\
        \textbf{50}  &  \textbf{13/13}  &  \textbf{12/13}  &  \textbf{12/13}  &  \textbf{12/13}  &  \textbf{13/13}  &  \textbf{9/13}  \\
        \bottomrule
    \end{tabular}
\end{table}

\begin{figure*}[ht]  % 用 [H] 需要 \usepackage{float}
    \centering
    \vspace{8pt}
    \includegraphics[width=0.95\linewidth]{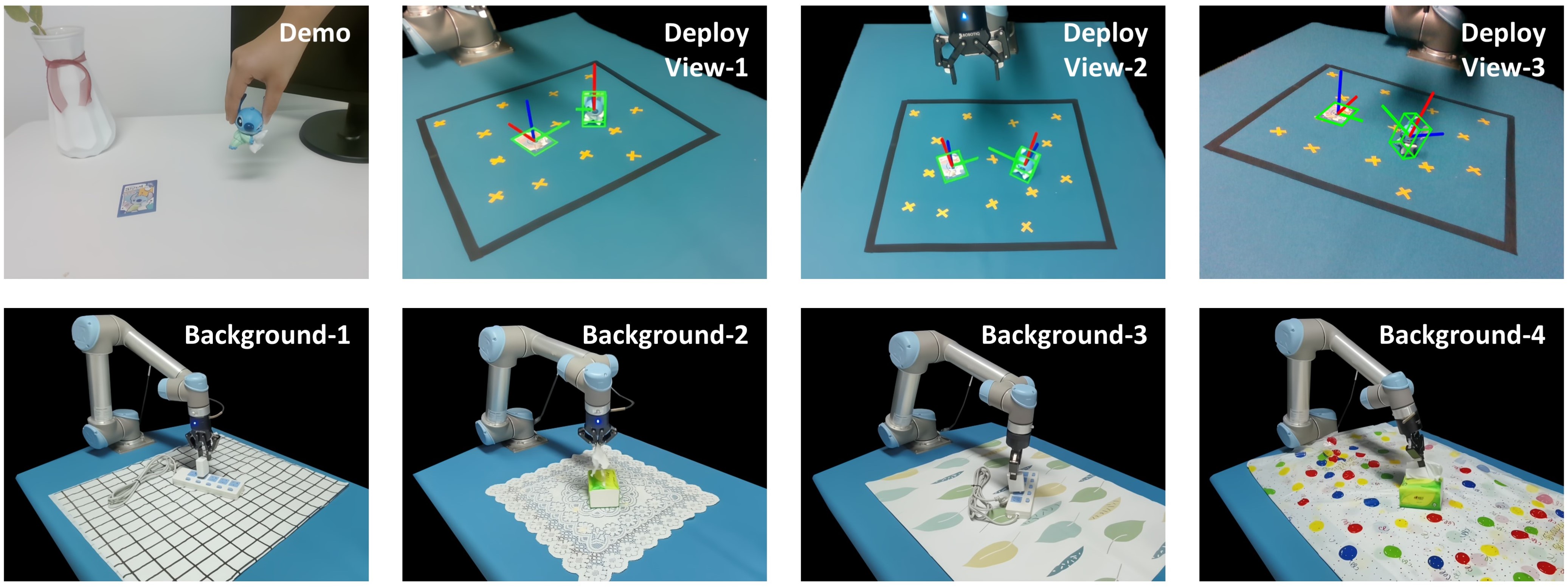}
    \caption{
    \textbf{Visual domain variation.} The top row shows the viewpoint of the human video demonstration, along with three different deployment viewpoints. The bottom row illustrates four types of background perturbations introduced during deployment.
    }
    \label{fig:view}
\end{figure*}

\begin{table*}[t]
    \centering
    \small
    \caption{Success rates under different visual domain variations. PG: position generalization; PoG: pose generalization.}
    \label{tab:PG_single}
    \renewcommand{\arraystretch}{1.2}
    \setlength{\tabcolsep}{3pt}

    \begin{tabular*}{\textwidth}{@{\extracolsep{\fill}}l *{10}{cc} cc}
        \toprule
        \multirow{2}{*}{\textbf{Task}}
        & \multicolumn{2}{c}{View-1}
        & \multicolumn{2}{c}{View-2}
        & \multicolumn{2}{c}{View-3}
        & \multicolumn{2}{c}{Background-1}
        & \multicolumn{2}{c}{Background-2}
        & \multicolumn{2}{c}{Background-3}
        & \multicolumn{2}{c}{Background-4}
        & \multicolumn{2}{c}{Average} \\
        \cmidrule(lr){2-3} \cmidrule(lr){4-5} \cmidrule(lr){6-7} \cmidrule(lr){8-9} 
        \cmidrule(lr){10-11} \cmidrule(lr){12-13} \cmidrule(lr){14-15} \cmidrule(lr){16-17}
        & \textbf{PG} & \textbf{PoG}
        & \textbf{PG} & \textbf{PoG}
        & \textbf{PG} & \textbf{PoG}
        & \textbf{PG} & \textbf{PoG}
        & \textbf{PG} & \textbf{PoG}
        & \textbf{PG} & \textbf{PoG}
        & \textbf{PG} & \textbf{PoG}
        & \textbf{PG} & \textbf{PoG} \\
        \midrule
        Cup-Coaster    & 5/5 & 5/5 & 5/5 & 5/5 & 5/5 & 4/5 & 5/5 & 5/5 & 5/5 & 5/5 & 5/5 & 5/5 & 5/5 & 5/5 & 35/35 & 34/35 \\
        Stitch-Card    & 5/5 & 5/5 & 4/5 & 4/5 & 5/5 & 4/5 & 5/5 & 5/5 & 5/5 & 5/5 & 5/5 & 5/5 & 5/5 & 4/5 & 34/35 & 32/35 \\
        Pour-Water     & 5/5 & 4/5 & 5/5 & 4/5 & 5/5 & 5/5 & 5/5 & 5/5 & 5/5 & 5/5 & 5/5 & 5/5 & 5/5 & 5/5 & 35/35 & 33/35 \\
        Pull-Drawer    & 5/5 & 4/5 & 5/5 & 5/5 & 5/5 & 4/5 & 5/5 & 5/5 & 5/5 & 5/5 & 4/5 & 5/5 & 5/5 & 5/5 & 34/35 & 33/35 \\
        Unplug-Charger & 5/5 & 5/5 & 4/5 & 5/5 & 5/5 & 5/5 & 5/5 & 5/5 & 5/5 & 4/5 & 5/5 & 5/5 & 4/5 & 5/5 & 33/35 & 34/35 \\
        Pull-Tissue    & 5/5 & 5/5 & 4/5 & 5/5 & 5/5 & 5/5 & 5/5 & 5/5 & 5/5 & 4/5 & 4/5 & 5/5 & 5/5 & 5/5 & 33/35 & 34/35 \\
        \bottomrule
    \end{tabular*}
    \vspace{-4pt}
\end{table*}
\subsubsection{Spatial Generalization}
% 为了验证我们的方法的空间泛化能力，我们在平面区域内设置了一个45x50cm的工作空间，用于均匀采样测试用例。由于我们仅依赖单挑人类视频数据，我们分别设置了位置泛化和位姿泛化实验，从简到难形成有效的泛化性对比。对于位置泛化性，我们在空间内随机选择了13个点，他们的相对距离分布充分覆盖了合法空间。而对于位姿泛化性，我们在这13个点的基础上进行了旋转随机化，针对每个位置我们随机选择了一个与所采数据不同的旋转。我们分别记录了仅用单条数据即不进行数据增广、增广到15条数据以及增广到50条数据的表现，结果如表？所示。
To evaluate the spatial generalization capabilities of our method, we define a workspace of 45 cm × 50 cm and randomly sample test cases within this region. \cwn{In this evaluation, the single human demonstration is used to generate an initial trajectory set, and we study how the size of this generated training set affects spatial generalization. }%Given that our approach relies on only a single human demonstration, %
We design two levels of spatial generalization: position generalization (PG) and pose generalization (PoG), progressing from simpler to more challenging settings. For position generalization, we randomly select 13 points whose spatial distribution covers diverse relative configurations across the workspace. For pose generalization, we apply random orientation perturbations at each of the 13 positions, generating novel configurations unseen in the demonstration trajectory. 

Since the relative motion in hand tasks is typically consistent, and the reference object pose can be estimated in real time, a single demonstration is often sufficient for spatial generalization. Therefore, we focus our spatial generalization evaluation on three object tasks. We report results under three training settings: one-shot (a single demonstration), 15-shot (augmented from one), and 50-shot (further augmented), as summarized in Table~\ref{tab:PG}. 
%Although trained on a single trajectory, TDV demonstrates a certain degree of spatial generalization. %
\cwn{When training uses the single generated trajectory without further expansion, TDV demonstrates a certain degree of spatial generalization.}
However, it tends to overfit to the single demonstration, consistently reverting to the initial pose seen during training at the beginning of each execution. Such behavior not only reduces efficiency but also increases the likelihood of collisions, as shown in Fig.~\ref{fig:1vs50}. Through efficient trajectory generation, we augment the dataset to 15 and 50 trajectories. As the number of training samples increases, TDV's spatial generalization improves significantly. With 15 trajectories, the coverage of the workspace remains relatively sparse, leading to invalid executions when the initial pose lies far from the augmented data distribution. When the dataset is expanded to 50 trajectories, the success rate approaches 100\%. These results confirm that TDV is capable of achieving spatial generalization from one-shot demonstration, thereby providing strong empirical evidence in support of Question~\ref{q:oneshot}.

\subsubsection{Visual Domain Generalization}
% 我们主要探究视觉域内中的背景泛化性和视角泛化以及cross-embodiedment能力。我们分别在与机械臂实验环境不同的背景下采集人类操作视频数据，并且在实验的时候我们选择三个与数据采集不同的相机视角，如图？所示。我们在每个相机视角下分别进行了5次位姿泛化性实验，其结果如表?所示。

For visual domain generalization, we primarily investigate three aspects: background variation, viewpoint variation, and cross-embodiment transfer. To evaluate background robustness, we collect human demonstration videos in environments with backgrounds distinct from the robot workspace, and introduce four types of background perturbations. For viewpoint variation, we conduct evaluations using three camera viewpoints that are distinct from those used during data collection, as illustrated in Figure~\ref{fig:view}. For each visual domain variation, we perform five trials with randomized poses, and the results are reported in Table~\ref{tab:PG_single}. Failure cases mainly occur when the camera adopts a steep top-down angle, making it difficult to localize small objects such as the Stitch toy, or when the object color closely resembles the background, resulting in tracking failure. 
However, such cases are relatively rare. Overall, our method demonstrates robustness to variations in camera viewpoints and background conditions, providing strong evidence in support of Question~\ref{q:domain}.

\section{CONCLUSIONS}
   % We propose a trajectory-based diffusion framework, TDV, which enables generalizable manipulation across different embodiments from one-shot human demonstrations. 我们通过充分的仿真和实物实验验证了轨迹这种中间表达方式能实现高效的数据生成，从而实现空间泛化性的操作。并且由于轨迹具有modality-agnostic的特性，因此表现出了强大的视觉域泛化性，在不同的相机视角和场景都有robust的表现。然而由于我们的方法很大程度上依赖6d pose estimation，因此对于形态复杂无法正常跟踪的物体以及柔性物体都无法正确操作。另外尽管数据生成能够维持原有轨迹的合法性，但是仍然无法解决避障问题，这是我们未来研究的一个方向。

    We propose a trajectory-based diffusion framework, TDV, which enables generalizable manipulation across different embodiments from one-shot human demonstrations. Through extensive simulation and real-world experiments, we validate that trajectories serve as an effective intermediate representation, allowing efficient data generation for spatial generalization. Furthermore, the modality-agnostic nature of trajectories contributes to strong robustness across visual domains, enabling consistent performance under varying camera viewpoints and background conditions. However, our approach relies heavily on accurate 6D pose estimation, which limits its applicability to objects with complex geometry or deformable structures that are difficult to track reliably.

\bibliographystyle{IEEEtran}
% \bibliography{refer}  % .bib
\bibliography{IEEEabrv,refer}
% \end{thebibliography}

\end{document}